\documentclass[letterpaper, 10 pt, conference]{ieeeconf}  
\IEEEoverridecommandlockouts
\usepackage{cite}
\usepackage{amsmath,amssymb,amsfonts}
\usepackage{graphicx}
\usepackage{array}
\usepackage{units}
\usepackage{makecell}
\usepackage{multirow}
\usepackage[nolist,nohyperlinks]{acronym}
\usepackage{textcomp}
\usepackage{breqn}
\usepackage{mathtools}
\usepackage{verbatim}

\usepackage{hyperref}
\newcommand{\citep}[1]{\cite{#1}}
\newcommand{\citet}[1]{\cite{#1}}
\usepackage[dvipsnames]{xcolor}
\usepackage[normalem]{ulem} 

\newcommand{\reviewchanges}[1]{{#1}} 
\newcommand{\finalsubmission}[1]{{#1}} 
\renewcommand{\sout}[1]{}

\begin{document}

\title{Experimental Comparison of Visual-Aided Odometry Methods for Rail Vehicles}
\author{Florian Tschopp$^1$, Thomas Schneider$^1$, Andrew W. Palmer$^2$, Navid Nourani-Vatani$^2$, Cesar Cadena$^1$,\\ Roland Siegwart$^1$, and Juan Nieto$^1$%
\thanks{$^1$Authors are members of the Autonomous Systems Lab, ETH Zurich, Switzerland; {\tt\small \{firstname.lastname\}@mavt.ethz.ch}}%
\thanks{$^2$Authors are with Siemens Mobility, Berlin, Germany; {\tt\small \{firstname.lastname\}@siemens.com}}%
\thanks{\copyright 2019 IEEE.  Personal use of this material is permitted.  Permission from IEEE must be obtained for all other uses, in any current or future media, including reprinting/republishing this material for advertising or promotional purposes, creating new collective works, for resale or redistribution to servers or lists, or reuse of any copyrighted component of this work in other works.}
}

\maketitle

\begin{abstract}
Today, \reviewchanges{rail vehicle} localization is based on infrastructure-side Balises (beacons) together with on-board odometry to determine whether a rail segment is occupied.
Such a coarse locking leads to a sub-optimal usage of the rail networks.
New railway standards propose the use of moving blocks centered around the rail vehicles to increase the capacity of the network.
However, this approach requires accurate and robust position and velocity estimation of all vehicles.
In this work, we investigate the applicability, challenges and limitations of current visual and visual-inertial motion \reviewchanges{estimation} frameworks for rail applications. 
An evaluation against RTK-GPS ground truth is performed on multiple datasets recorded in industrial, sub-urban, and forest environments.
Our results show that stereo visual-inertial odometry has a great potential to provide a precise motion estimation because of its complementing sensor modalities and shows superior performance in challenging situations compared to other frameworks.
\end{abstract}

\section{Introduction}
%
In recent years, the need for public transportation has risen dramatically. 
Rail transportation alone has increased by over $\unit[60]{\%}$ in the last 16 years in Switzerland \citep{BundesamtfurStatistik2017}. 
However, current infrastructure is reaching its capacity limits. 
To keep up with this growth, there is a  need to improve the system efficiency.

In train applications, a crucial part of the current infrastructure is the traffic control system.
Most of current rail control systems divide the railroad tracks into so-called blocks \citep{Stanley2011}.
The block size is determined by the worst case braking distance of every vehicle that is likely to operate on this track. 
Vehicle localization and interlocking of the blocks is performed using infrastructure-side beacons. 
Such a fixed block strategy results in very conservative interlocking and thus, decreases the overall efficiency of the system.

The new \ac{etcs} Level~3 aims to replace the fixed blocks with moving blocks centered around the vehicle.
This concept has the potential of increasing the capacity of train networks by \finalsubmission{a factor of} $\unit[190]{\%}$ to $\unit[500]{\%}$~\citep{Williams2016}.
Furthermore, fixed track-side sensing infrastructure (e.g. axle-counters, Balises) may be replaced with on-board sensors, leading to a more cost-effective solution in the long-run.
Even with the vast amount of research in related applications (e.g. autonomous cars), the success of \ac{etcs} Level 3 is subject to the development of new algorithms that are able to precisely and reliably estimate both the position and velocity of all rail vehicles \reviewchanges{\cite{Beugin2012Simulation-basedLocalization,Marais2017ASignaling}}. 
\begin{figure}
\centering
\includegraphics[width=0.8\columnwidth]{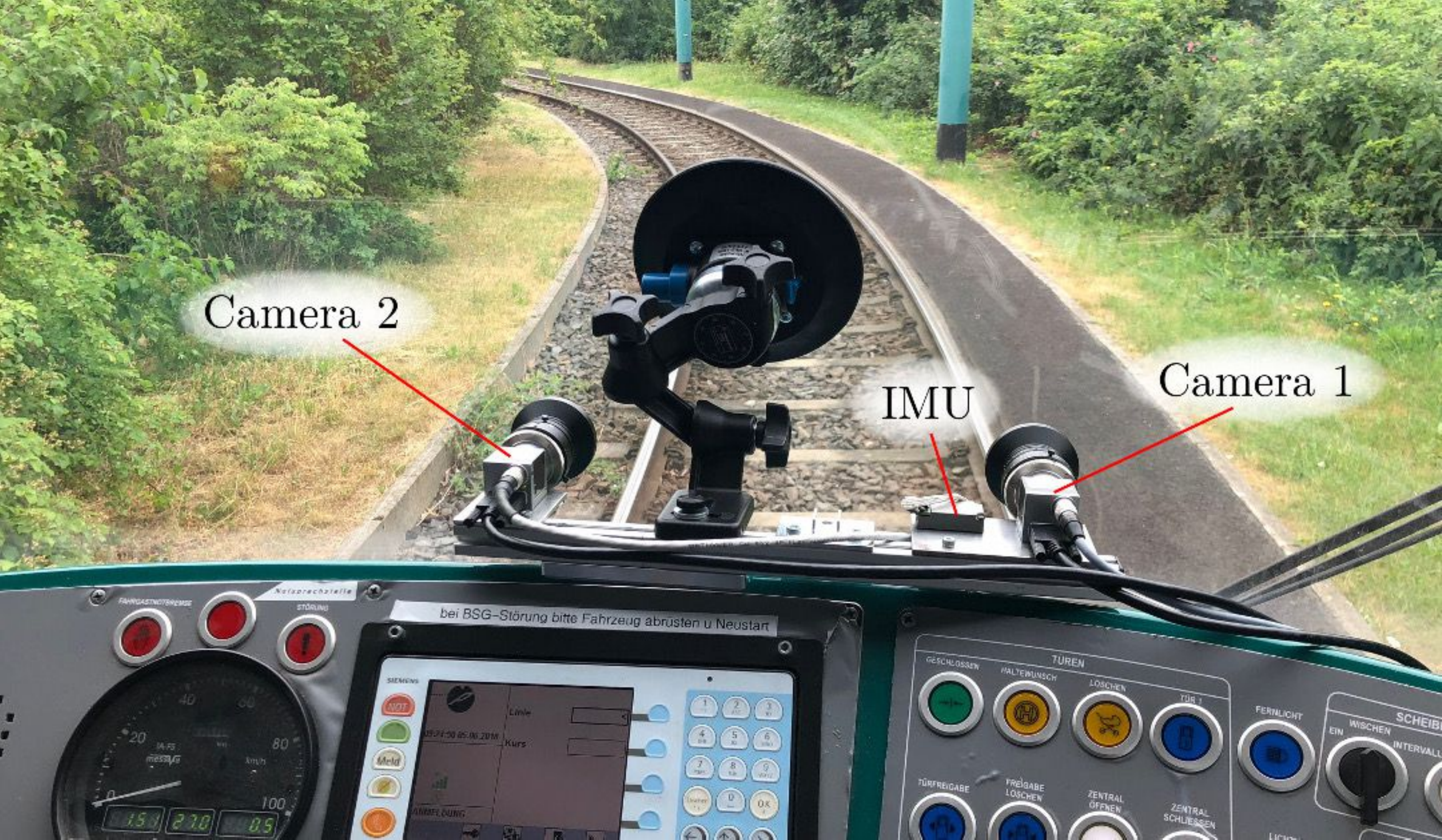}
\caption{Datasets recorded with a custom sensor setup for visual-aided odometry in sub-urban and industrial environments are used for evaluation of popular visual-aided odometry frameworks for rail applications. We show that high accuracy motion estimation can be achieved using stereo vision. Furthermore, incorporating inertial measurements increases accuracy and robustness.} \label{fig:31cm}
\end{figure}

In rail applications only few restrictions exist in regard to weight and power consumption of the localization solution.
Therefore, one is pretty open in choosing suitable sensor modalities and estimation algorithms.
Current research in train localization mainly focuses \reviewchanges{on} the fusion of \ac{gnss} with inertial measurements coupled with infrastructure-side beacons.
In safety critical application such as train localization, a high level of reliability can only be achieved using redundant and complementary sensors.

Recently, the robotics and computer vision communities have reported visual motion estimation and localization systems with an impressive accuracy and robustness~\citep{Bloesch2015,Mur-Artal2015ORB-SLAM:System,Engel2018DirectOdometry,Qin2018VINS-Mono:Estimator,Burki2016}. 
We believe that synchronized visual and inertial sensors have the right properties to be an ideal extension to the currently used sensor modalities.
\reviewchanges{
A continuous global localization is often not feasible using vision sensors due to ambiguous environments or drastic appearance changes.
However, combining incremental odometry information with localization to reduce drift accumulated by the odomertry method can provide a continuous and high-accuracy pose estimation.
}
For this reason, \reviewchanges{as a first step towards such a system,} we want to investigate current state-of-the-art visual(-inertial) motion estimation frameworks for their applicability on train applications.
\reviewchanges{The main challenges include high speeds, constrained motion \reviewchanges{leading to potential observability issues of IMU biases \cite{Martinelli2014Closed-formMotion}}, challenging lighting conditions and highly repetitive patterns on the ground.}

Our contribution consists of the first evaluation of popular generic visual-aided odometry frameworks for rail applications.
We use a \ac{rtk} \ac{gps} device as ground truth, in order to evaluate the pose estimates on datasets recorded on two trajectories in an industrial, sub-urban and forest environment.
Furthermore, we identify, investigate, and discuss specific challenges of current methods.

\section{Related Work}
Motion estimation is the backbone of many established small autonomous ground robots and has had an increasing presence due to the rise of autonomous driving. 
Current solutions in commercial products rely on the use of  sensor fusion of \ac{gnss} (in outdoor scenarios), wheel odometry, and additional sensors such as Light Detection And Ranging devices (LiDARs) or cameras, e.g. \citep{Hitz2018, Rendall2018, SA2018}. 

In contrast to generic ground robots, trains have a distinct feature: their motion is constrained to the rail network. 
This paper studies advantages and disadvantages this constraint implies to the motion estimation performance compared to more generic approaches. 

\subsection{Rail vehicle odometry and localization}
The current research goal is to increase the accuracy and robustness of motion estimation and localization.
The approaches are split into improving the in-place infrastructure (e.g. track-side Balises, odometer and speed sensors) or investigating new sensor modalities.

Mourillas and Poncet \citet{Murillas2016} and Palmer and Nourani-Vatani \citet{Palmer2018} describe measures on how to increase the robustness and reliability of the currently used on-board odometry measurements using wheel encoders and ground speed radars. 
Recent works also include using machine learning approaches such as \acp{lssvm} to improve the localization accuracy \citep{Cheng2017}. 

To decrease the dependency on track-side infrastructure, new sensor modality research is highly focused on the use of an \ac{imu} together with a tachometer \citep{Allotta2015} or \ac{gnss}. 
Otegui et al. \citet{Otegui2017} summarizes many works fusing \ac{imu} and \ac{gps} signals employing an \ac{ekf} or a \ac{pf}. 

To further improve accuracy, Hasberg et al. \citet{Hasberg2012} and Heirich et al. \citet{Heirich2013a} introduce \ac{slam} for path-constrained motion.
Fusing \ac{gnss} with \ac{imu} measurements and a path constraint, high accuracies in position retrieval ($<\unit[1]{m}$) and map building with a precision of around $\unit[10]{cm}$ are presented \citep{Hasberg2012}.

All these existing methods rely heavily on \ac{gps}, \ac{imu} and wheel encoders / ground speed radars, each of which have their own failure cases.
For instance \ac{gps} has denied areas, suffers from multi-path effects near structures and is easy to jam, \ac{imu} bias may become unobservable \citep{Martinelli2014Closed-formMotion,Du2017}, encoders suffer from wheel slippage or mis-calibration \citep{Palmer2018} and radars have problems with reflectance off the ground \citep{Murillas2016,Spindler2016}. 
To improve the robustness and achieve high safety levels through redundancies, additional sensing modalities such as cameras will be critical.

Wohlfeil \citet{Wohlfeil2011} performs visual track recognition based on edge extraction with a known track gauge width. 
\reviewchanges{\sout{The method was shown to work reliably, but was not tested at night, in tunnels or during heavy snowfall, rainfall or fog; conditions that may render this approach unusable.}}
Failure cases were\reviewchanges{\sout{also}} observed where switches were missed or confused, especially in challenging lighting conditions (e.g. bright sky). 
Furthermore, a continuous position estimate is not provided, only the direction of travel after a switch. 
Bah et al. \citet{Bah2009} present a pure vision-based odometry method for trains by transforming a front facing camera image to a birds-eye view of the track and finding correspondences on two consecutive frames. 
This method might fail with low-texture or repetitive grounds which often occur in train environments.

The mentioned visual-aided odometry and localization algorithms are\reviewchanges{\sout{ highly specific for the rail application and do not provide the desired level of accuracy.
Also, imposing such constraints to the problem might reduce the level of redundancy.} not directly suitable for continuous motion estimation as they only provide information near switches \cite{Wohlfeil2011} or require manual data association \cite{Bah2009}.
Furthermore, by not considering specific constraints, the visual odometry can later be fused with this information to get even more reliable and accurate pose estimation, to detect when a method is failing or to detect changes in the expected environment \cite{Altendorfer2010SensorSystems}.}
The goal of this paper is to study the performance of generic visual-aided ego-motion estimation for the rail application.

\subsection{Generic visual-aided ego-motion estimation}

State-of-the-art approaches in odometry estimation and \ac{slam} using vision sensors can be classified into filter approaches (mostly \ac{ekf}), where typically all the past robot poses are marginalized and only the most recent pose is kept in the state \citep{Bloesch2015,Davison2007}, and sliding-window approaches, where multiple poses are kept in an optimization \citep{Mur-Artal2015ORB-SLAM:System,Engel2018DirectOdometry}. 

Sliding-window based approaches are studied in detail by Strasdat et al. \citet{Strasdat2010}, proving to outperform filter-based approaches when employing the same computational resources.
Furthermore, sliding-window based approaches are very flexible for incorporating measurements from different sensing modalities with different propagation models.
To keep the computational costs within hardware limits, these schemes typically limit the state to within a sliding window. 
Efforts to unify both approaches are presented by Bloesch et al. \citet{Bloesch2017TheRobots}.

The most prominent visual odometry methods are probably \textit{ORB-SLAM} \citep{Mur-Artal2015ORB-SLAM:System} and \textit{\ac{dso}} \citep{Engel2018DirectOdometry}.
\textit{ORB-SLAM} extracts and tracks keypoint features in an image stream while \textit{\ac{dso}} uses a direct approach based on image gradients. 
These methods cannot recover the metric scale of the map, which is critical in the given application. 

One prominent method to recover the metric scale is adding an \ac{imu}. 
Extending the previous works in \citep{vonStumberg2018DirectMarginalization} and \citep{Mur-Artal2017Visual-InertialReuse} respectively, the scale can be observed by incorporating inertial measurements, often referred to as \ac{vio}. 
However, depending on the performed motion, the \ac{imu} biases are not fully observable \citep{vonStumberg2018DirectMarginalization,Du2017,Martinelli2014Closed-formMotion} resulting in errors in scale estimation. 

Another method to recover the scale is stereo vision shown in \citep{Mur-Artal2017} and \citep{Wang2017StereoCameras}.
Leutenegger et al. \citet{Leutenegger2015} proposes the combination of stereo vision and \ac{imu} measurements, resulting in a reliable feature-based sparse map and accurate camera pose retrieval. 

\reviewchanges{In automotive applications, many of the challenges such as high velocities and constraint motion are similar to the rail domain.
There, odometry is often solved by using wheel encoders \cite{Schwesinger2016AutomatedE-mobility, Merriaux2014WheelMap} as they do not suffer from high slip as in rail applications.
Furthermore, stereo vision \cite{Churchill2012ExperienceImplementation} and monocular \ac{vo} with learned depth \cite{Barnes2017DrivenEnvironments} have also been used successfully for ego motion estimation.
}
A multitude of state-of-the\reviewchanges{-}art stereo-visual odometry frameworks are tested \reviewchanges{for automotive applications} in the visual odometry part of the KITTI Vision Benchmark Suite \citep{Geiger2012}. 
One popular and well performing open-source pipeline is \textit{ORB-SLAM2} \citep{Mur-Artal2017}.
\reviewchanges{
Unfortunately, the KITTI dataset does not include synchronized \ac{imu} measurements and therefore does not allow in-depth insights into the benefits inertial data could provide.
Finally, the scale of the motion can also be retrieved by exploiting non-holonomic constraints \cite{Scaramuzza2009AbsoluteConstraints} which, however, relies on frequent turns.

In contrast to the mentioned approaches for automotive applications, this paper aims to investigate the benefit inertial data can provide for motion estimation in the rail domain and compares it to already successfully deployed stereo visual odometry.} 

\section{Visual-aided odometry pipelines} \label{sec:method}
\begin{table*}[t]
\centering
\caption{Overview of visual-aided odometry approaches.} \label{tab:estimators_overview}
\begin{tabular}{l|ccc|ccc|p{6cm}}
&\multicolumn{3}{c|}{Estimator type} & \multicolumn{3}{c|}{Sensor measurements}&Comment \\ 
& \ac{ekf} & Sliding-window & Batch & Monocular & Stereo & \ac{imu} \\ \hline
ROVIO \citep{Bloesch2015} & $\times$ & &  & $\times$ & & $\times$ & Light-weight EKF based \ac{vio} using patch tracking.\\
VINS-Mono \citep{Qin2018VINS-Mono:Estimator} &  &$\times$ & & $\times$ & & $\times$& Tightly coupled indirect monocular VI fusion. \\
Batch optimization \citep{Schneider2017} &  & & $\times$& $\times$ & & $\times$& Offline global batch VI bundle-adjustment. \\
ORB-SLAM2 \citep{Mur-Artal2017} & &$\times$  & &  &$\times$ &&Indirect stereo visual SLAM framework.\\
OKVIS \citep{Leutenegger2015} &  &$\times$ & &  & $\times$ & $\times$ & Keyframe based tight coupling of stereo VI fusion.\\
Stereo-SWE \citep{Hinzmann2016MonocularOptimization} & &$\times$ & &   &\reviewchanges{$\times$} & \reviewchanges{$\times$} & Tightly coupled VI fusion using depth as independent measurement. \\
\end{tabular}
\end{table*}
In order to evaluate the performance of visual-aided ego-motion estimation for rail applications, we made a selection of the most promising available pipelines summarized in \autoref{tab:estimators_overview}.
\reviewchanges{\sout{Main challenges include high speeds, constrained motion, challenging lighting conditions and highly repetitive patterns on the ground.}}

\subsection{Visual-inertial odometry algorithms}
The goal of \ac{vio} is to increase robustness and observe the scale of the motion using inertial measurements. 
Advantages of \ac{vio} are gravity aligned maps and complementing sensor modalities.
One disadvantage is the dependency on specific motion patterns in order to make the biases observable.

In this paper, the following state-of-the-art algorithms are introduced and further evaluated.
\subsubsection{ROVIO}
In \citep{Bloesch2015}, a light-weight visual-inertial odometry algorithm based on an \ac{ekf} is presented. 
It shows high robustness even in very fast rotational motions. 
\textit{ROVIO} directly uses pixel intensity errors on patches and can therefore be considered a direct method.

\subsubsection{VINS-Mono}
Qin et al. \citet{Qin2018VINS-Mono:Estimator} proposes a \ac{vio} algorithm based on indirect tightly coupled non-linear optimization.
Compared to a stereo visual-inertial pipeline like \textit{OKVIS} \reviewchanges{\cite{Leutenegger2015}} (see Section \ref{sec:combined}) which can also deal with stereo cameras, \textit{VINS-Mono} is specifically designed for monocular \ac{vio} with main differences in the initialization procedure.
Furthermore, \textit{VINS-Mono} reports slightly better accuracy results in \ac{uav} applications \citep{Delmerico2018ARobots} compared to\reviewchanges{\sout{ stereo visual inertial approaches} \textit{OKVIS} when used in monocular mode}.

\subsubsection{Batch optimization}
Using \textit{ROVIO} \citep{Bloesch2015} as an estimator, \textit{ROVIOLI} is an online front-end to build maps in the \textit{maplab} \citep{Schneider2017} format. 
The created map can be post-processed using \textit{maplab} batch bundle-adjustment to reduce drift and correct for linearization errors.

\subsection{Stereo visual odometry algorithms}
In addition to using inertial measurements, the metric scale of the motion can also be immediately retrieved using depth measurements of a stereo camera pair.
In contrast to \ac{vio}, stereo-visual odometry does not require specific motions.
However, as the method is purely visual, it will fail whenever the visual system faces challenges in tracking landmarks.
\subsubsection{ORB-SLAM2}
Mur-Artal and Tardos \citet{Mur-Artal2017} provide a complete visual \ac{slam} system for monocular, stereo or RGB-D cameras called \textit{ORB-SLAM2}.
The odometry front-end of the system is based on matching ORB features. 
The optimization is performed on a pose graph only containing the most relevant keyframes.
Stereo constraints are incorporated in the cost function by projecting the landmarks with successful stereo matches to an augmented keypoint measurement including the coordinates of both cameras.

\subsection{Stereo visual-inertial algorithms} \label{sec:combined}
In order to compensate for failure cases of the previously mentioned approaches, stereo vision and inertial measurements can be combined into a unified framework.

\subsubsection{OKVIS}
Leutenegger et al. \citet{Leutenegger2015} introduce tight-coupling of inertial and indirect visual measurements in a keyframe based approach optimizing inertial and re-projection errors in a sliding-window.
In addition to the previously mentioned algorithms, \textit{OKVIS} is able to deal with both stereo cameras by fusing landmarks visible in both frames and inertial data by using pre-integrated factors \citep{Lupton2012Visual-inertial-aidedConditions}.

\subsubsection{Stereo-SWE}
Fusing landmarks, such as in \textit{OKVIS}, can result in problems if the stereo matches contain wrong matches or outliers.
Even if a robust cost function could avoid taking them into account, all additional information this landmark could provide is lost after a wrong merge.
Alternatively, stereo matches could also be used as additional independent measurements for each landmark observation instead of fusing the landmarks.

Due to the lack of an available implementation for this approach, we extended the visual-inertial \textit{\ac{swe}} presented by Hinzmann et al. \citet{Hinzmann2016MonocularOptimization}.
In addition to the mentioned re-projection error and inertial error, a weighted depth error is introduced for each landmark observation with stereo matches.
The depth error is the difference of the measured depth obtained by triangulating the stereo matches and the depth of the corresponding landmark projected to the camera. 
Inspired by \citep{Mur-Artal2017}, depth error uncertainties are scaled by their distance to the cameras and only considered up to a certain distance relative to the baseline between the cameras.

\section{Experimental evaluation} 
In this section, the experimental evaluation of the mentioned algorithms is shown\footnotemark[1].
\footnotetext[1]{\reviewchanges{All evaluations were performed on an Intel Xeon E3v5, $\unit[48]{GB}$ RAM laptop but not in real-time ($\unit[2-4]{fps}$).}}

We start with describing the datasets, explain recoding and evaluation procedure, and show the results and some in-depth analysis.
\subsection{Datasets}
\begin{figure}
\centering
\includegraphics[width=0.65\columnwidth, trim={0 2.5cm 0 2.5cm}, clip]{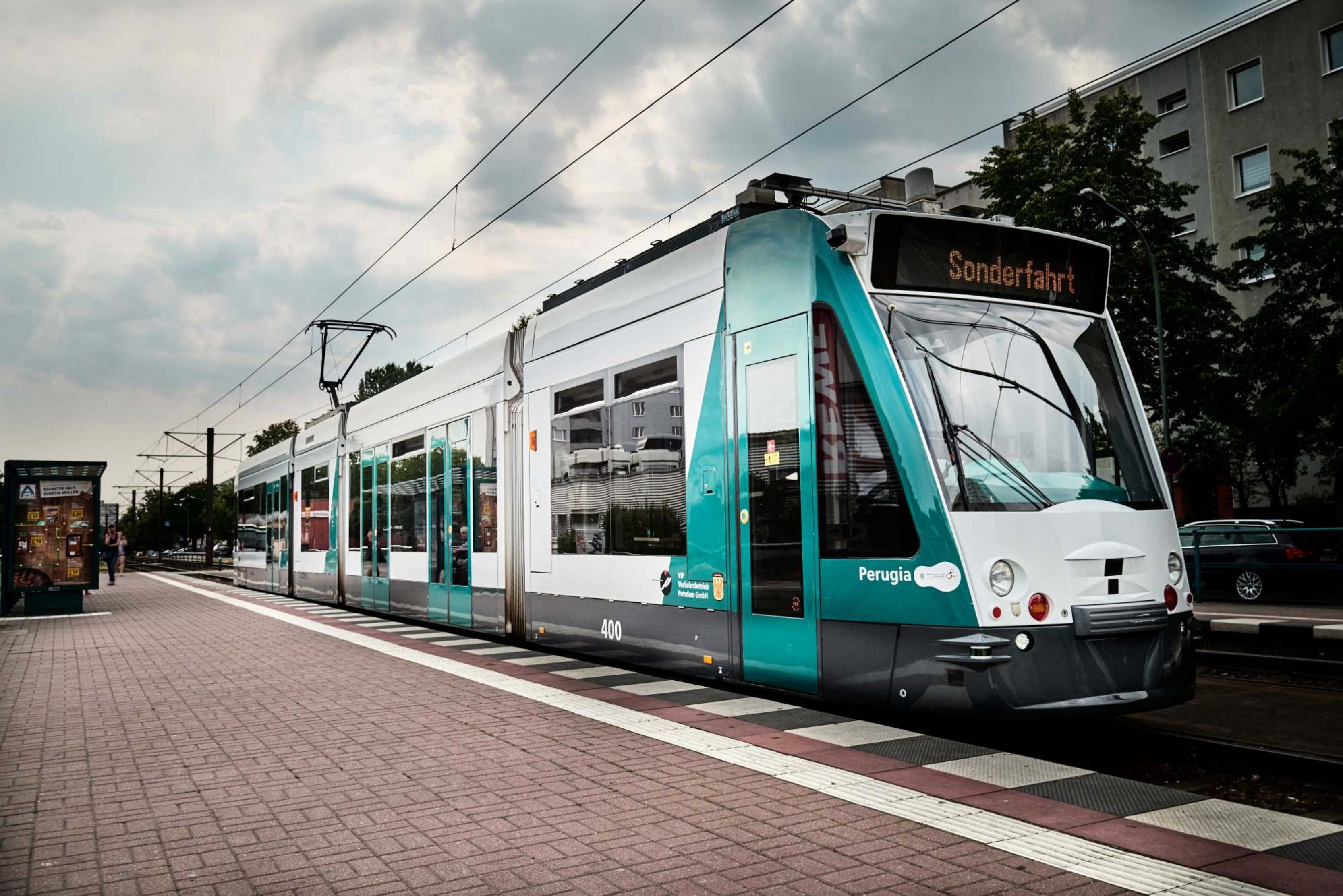}
\caption{Customized Siemens Combino test vehicle \citep{SiemensMobilityGmbH2018SiemensTram} for data collection in Potsdam, Germany.}\label{fig:combino}
\end{figure}
\begin{figure} 
\centering
\includegraphics[width=0.85\columnwidth]{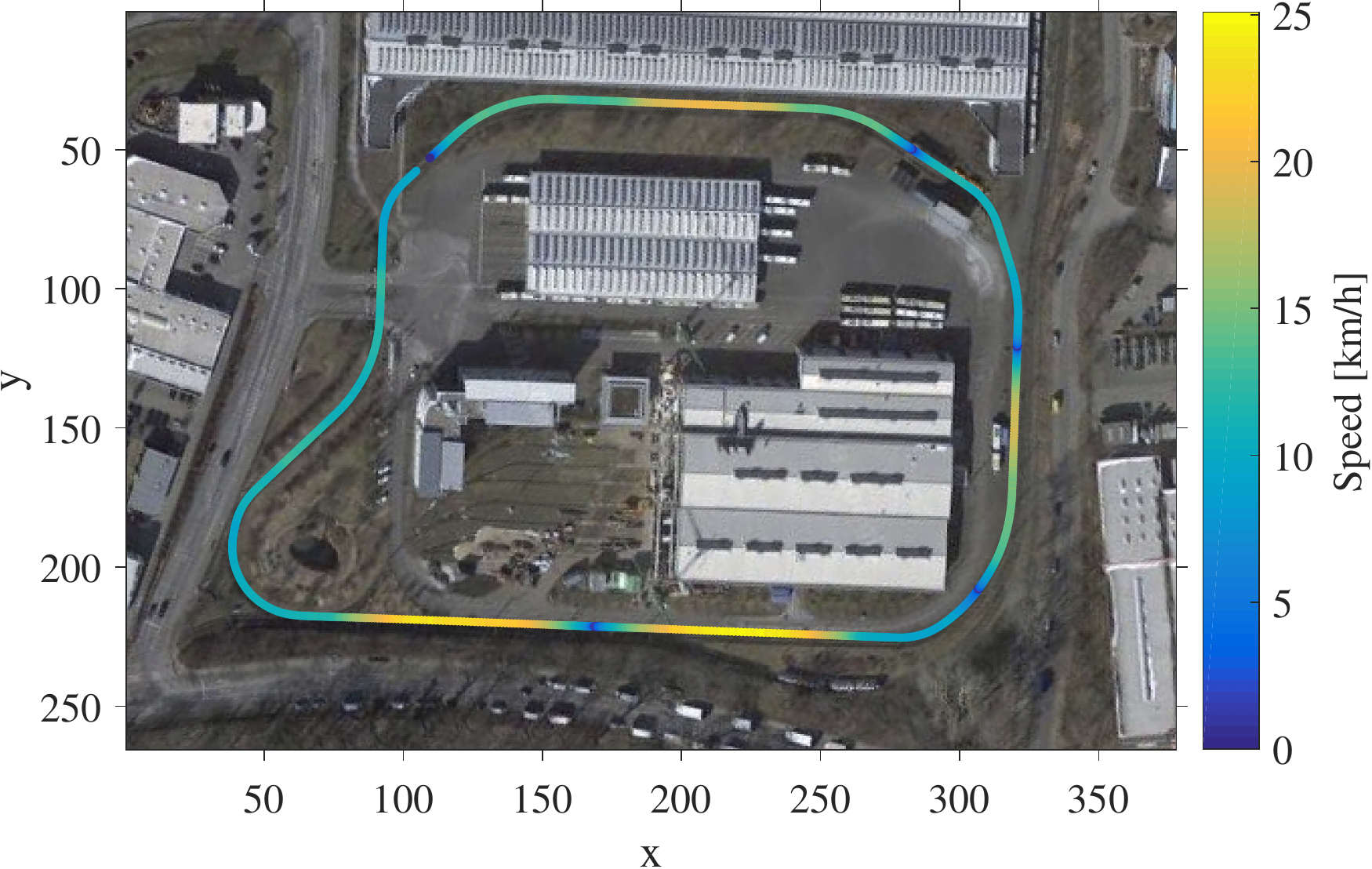}
\caption{\texttt{Trajectory1} around the depot in an industrial environment with speeds up to $\unitfrac[25.5]{km}{h}$.} \label{fig:betriebshof}
\end{figure}

Due to the lack of suitable available datasets, the estimators are tested on data recorded in Potsdam, Germany on a Siemens Combino tram (see \autoref{fig:combino}), which is customized for autonomous driving tests. 
\texttt{Trajectory1} is a short $\unit[780]{m}$ low-velocity (up to $\unitfrac[25.5]{km}{h}$) track around the depot in an industrial environment and close-by structures as shown in \autoref{fig:betriebshof}. 
\texttt{Trajectory2} is along a public tram-line about $\unit[2900]{m}$ long with speeds up to $\unitfrac[52.4]{km}{h}$ \reviewchanges{representing a real-life scenario}.
This trajectory includes rural, sub-urban, urban, and woody environments.
\reviewchanges{The datasets were captured on a sunny day in August 2018 as dealing with extreme conditions for visual sensing (rain, fog, nighttime) is beyond the scope of this paper.}
\begin{figure} 
\centering
\includegraphics[width=\columnwidth]{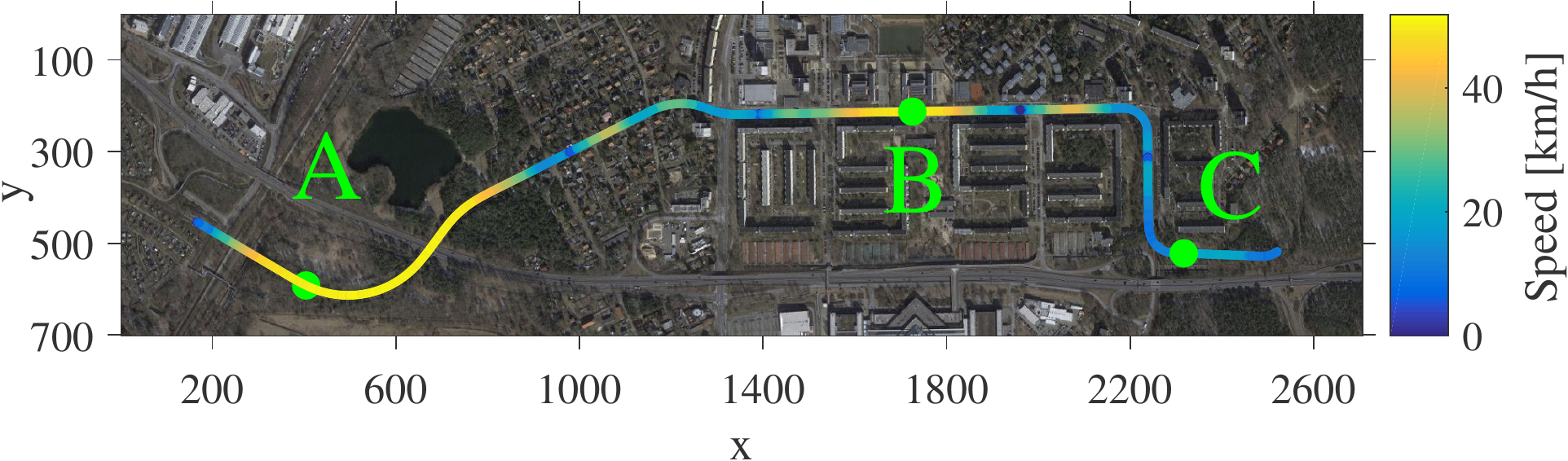}
\caption{\texttt{Trajectory2} following a public tram-line in Potsdam, Germany featuring a rural, sub-urban, urban, and woody environment and speeds up to $\unitfrac[52.4]{km}{h}$. The green letters indicate challenging scenarios discussed in Section \ref{sec:discussion_and_results}.} \label{fig:outside}
\end{figure}
\subsection{Hardware setup}
\subsubsection{\ac{vio} setup} \label{sec:vi-setup}
For the data collections, we deployed a custom-built stereo visual-inertial sensor which is synchronizing all measurements in hardware similar as in \citep{Nikolic2014}.
To feature higher accuracy, exposure time compensation is utilized\citet{Furgale2013UnifiedSystems}.

The sensor consists of two global-shutter cameras arranged in a fronto-parallel stereo setup and a compact, precision six degrees of freedom \ac{imu}. 
The camera was selected to provide a high frame-rate to be able to get a reasonable number of frames per displacement, even at higher speeds, and also to feature a high dynamic range to deal with the challenging lighting conditions. 
The \ac{imu} was chosen to feature low noise values and also to support temperature calibration, as direct sunlight can highly change the temperature of the sensors. 
The sensor specifications are summarized in \autoref{tab:vi-sensor}. 
\begin{table}
\caption{Sensor specifications deployed for data collection.}\label{tab:vi-sensor}
\begin{tabular}{m{1cm}|m{1.8cm}|m{4.8cm}} 
Device&Type&Specification \\ \hline \hline
Camera& Basler acA1920-155uc&\makecell[l]{Frame-rate $\unit[20]{fps}$\footnotemark[2], \\Resolution $1920 \times 1200$, \\Dynamic range $\unit[73]{dB}$}\\ \hline
Lens\reviewchanges{\sout{e}}& Edmund Optics&\makecell[l]{Focal length $\unit[8]{mm} \approx 70 \deg$ opening \\ angle; Aperture $f/5.6$}\\ \hline
IMU& ADIS16445& \makecell[l]{Temperature calibrated, $\unit[300]{Hz}$, \\ $\unitfrac[\pm250]{\deg}{s}$, $\unitfrac[\pm 49]{m}{s^2}$}\\ 
\end{tabular}
\end{table}
\footnotetext[2]{The hardware is able to capture up to $\unit[164]{fps}$.}
In order to investigate the influence of the baseline distance of the stereo setup, data was collected with three baselines of $\unit[31]{cm}$, $\unit[71]{cm}$ and $\unit[120]{cm}$.
Those baselines were chosen to have a wide variety from baselines common in automotive applications up to \reviewchanges{\sout{such from}those used in} fixed wing \acp{uav}.
\autoref{fig:31cm} shows the deployed sensor in a $\unit[31]{cm}$ baseline configuration mounted behind the windshield inside the front cabin of the test vehicle.

\subsubsection{Calibration}
Sensor intrinsic and extrinsic calibration was performed using the \textit{Kalibr} toolbox \citep{Furgale2013UnifiedSystems}. 
The transformation of the \ac{imu} with respect to the master camera (camera~1) is constant and calibrated in a lab environment. 
The transformation between the two cameras is then determined separately in-situ \reviewchanges{using a $7\times5$ checkerboard with tile sizes of $\unit[10.8]{cm}$. 
Even though a larger calibration target might be beneficial to enable mutual observations, for the larger baselines we needed a board of $\unit[1.5\times 2]{m}$ \finalsubmission{and} $\unit[3\times 4]{m}$, \finalsubmission{respectively,} which \finalsubmission{\sout{is}are} \finalsubmission{more} difficult to manufacture and handle \finalsubmission{and were not available during data collection}. }

\subsubsection{Ground truth}
Ground truth data is acquired using the high-precision \ac{rtk} \ac{gnss} device OTXS RT3005G. Typical accuracies of $\unit[0.05]{m}$ and $\unit[0.1]{\deg}$ are possible after post-processing.

\subsection{Evaluation}
The main metrics used in this paper to evaluate the performance of visual-aided odometry pipelines are incremental distance and heading errors. 

We use the segment-based approach introduced by Nourani-Vatani and Borges \citet{Nourani-Vatani2011} to deal with unbound errors in odometry \citet{Kelly}. 
Thereby, the trajectory is divided in segments. 
Two different segment lengths $\unit[10]{m}$ and $\unit[50]{m}$ are evaluated to test the evolution of errors.
Each segment is aligned with the corresponding ground truth trajectory segments using the first $\unit[10]{\%}$ of the segment. 
The distance error then corresponds to the distance between the end-points of the segments. 
The heading error is the difference in heading estimation between the two segment ends.

\subsection{Results and discussion} \label{sec:discussion_and_results}
\begin{table*}[t]
\centering
\caption{Result overview of \reviewchanges{\sout{median}estimation} errors (distance in $\unit{\%}$ / heading in $\unitfrac{\deg}{m}$) for the $\unit[31]{cm}$ baseline configuration. The best pipeline in the respective scenario \reviewchanges{and error metric (median or \ac{rmse})} is emphasized in bold letters. $^3$~Contains resets of the estimator.} \label{tab:result_estimator_performance}
\begin{tabular}{c|lr||cc|cc}
&Trajectory&& \multicolumn{2}{c|}{\texttt{Trajectory1}} & \multicolumn{2}{c}{\texttt{Trajectory2}}\\
&Segment length&& $\unit[10]{m}$ & $\unit[50]{m}$ & $\unit[10]{m}$ & $\unit[50]{m}$\\
\hline \hline
\multirow{6}{*}{Visual-inertial}  &\multirow{2}{*}{ROVIO \citep{Bloesch2015}$^3$} & Median & $66.570 / 0.0490$ & $67.723 / 0.0578 $ &  $75.292 / 0.0269 $ & $75.149 / 0.0210$\\
&&\reviewchanges{RMSE}& \reviewchanges{$74.620 / 0.1471$} & \reviewchanges{$67.468 / 0.1119 $} &  \reviewchanges{$77.297 / 0.0632 $} & \reviewchanges{$74.035 / 0.0511$}\\
&\multirow{2}{*}{VINS-Mono \citep{Qin2018VINS-Mono:Estimator}$^3$}& Median & $5.060 \,/\, 0.1033 $ & $10.589 \,/\, 0.4093 $ &  $43.552 \,/\,0.0408 $ &  $45.339 \,/\,0.0525 $\\
&&\reviewchanges{RMSE}& \reviewchanges{$783.40 \,/\, 0.5966 $} & \reviewchanges{$250.9 \,/\, 0.5741 $} &  \reviewchanges{$685.78 \,/\,0.2412 $} &  \reviewchanges{$274.54 \,/\,0.1805 $}\\
&\multirow{2}{*}{Batch optimization \citep{Schneider2017}} & Median & $7.092 \,/\, 0.0322 $ & $2.899 \,/\,0.0066 $ & $12.361 \,/\, 0.0153 $&  $4.239 \,/\, 0.0084 $\\
&&\reviewchanges{RMSE}& \reviewchanges{$9.050 \,/\, 0.0685 $} & \reviewchanges{$4.396 \,/\,0.0111 $} & \reviewchanges{$17.336 \,/\, 0.0302 $}&  \reviewchanges{$10.90 \,/\, 0.0143 $}\\ \hline
\multirow{2}{*}{Stereo visual}&\multirow{2}{*}{ORB-SLAM2 \citep{Mur-Artal2017}} & Median & $\mathbf{2.138} \,/\, \mathbf{0.0204} $ & $3.054 \,/\, 0.0093 $ &  $1.786 \,/\, 0.0078 $ & $1.829 \,/\, \mathbf{0.0033} $\\ 
&&\reviewchanges{RMSE}& \reviewchanges{$3.751 \,/\, 0.0605 $} & \reviewchanges{$5.026 \,/\, 0.0436 $ }&  \reviewchanges{$4.526 \,/\, 0.0126 $ }& \reviewchanges{$3.956 \,/\, 0.0073 $}\\ \hline
\multirow{4}{*}{\shortstack{Stereo visual-\\
 inertial}}&\multirow{2}{*}{OKVIS \citep{Leutenegger2015}} &Median &$2.152 \,/\, 0.0219 $  & $\mathbf{2.850} \,/\, \mathbf{0.0070} $ &$\mathbf{1.428} \,/\, \mathbf{0.0074} $  &$\mathbf{1.110} \,/\, 0.0038 $ \\
&&\reviewchanges{RMSE} & \reviewchanges{$\mathbf{3.732} \,/\, 0.0336 $}  &\reviewchanges{ $\mathbf{4.295} \,/\, \mathbf{0.0103} $} &\reviewchanges{$\mathbf{3.361} \,/\, \mathbf{0.0116} $}  &\reviewchanges{$\mathbf{2.907} \,/\, \mathbf{0.0055} $} \\
&\multirow{2}{*}{Stereo-SWE \citep{Hinzmann2016MonocularOptimization}}& Median& $2.845 \,/\, 0.0249$ & $4.029\,/\, 0.0128$ & $3.710 \,/\, 0.0099$ & $3.840\,/\, 0.0087$\\
&&\reviewchanges{RMSE}& \reviewchanges{$7.640 \,/\, \mathbf{0.0332}$} & \reviewchanges{$5.742\,/\, 0.0113$} & \reviewchanges{$5.552 \,/\, 0.0151$} & \reviewchanges{$4.998\,/\, 0.0116$}\\
\end{tabular}
\end{table*}
All ego-motion estimation pipelines investigated in Section \ref{sec:method} are tested on both trajectories using the different baselines. 
\reviewchanges{\sout{For a fair comparison, the state-of-the-art pipelines only experienced minor modifications and are used out-of-the-box.} To enable a high level of comparability, the state-of-the-art pipelines are used out-of-the-box with only minor tuning.
}
As pure odometry is under investigation here, all loop closing capabilities are disabled.

\begin{figure} 
\centering
\includegraphics[width=0.75\columnwidth]{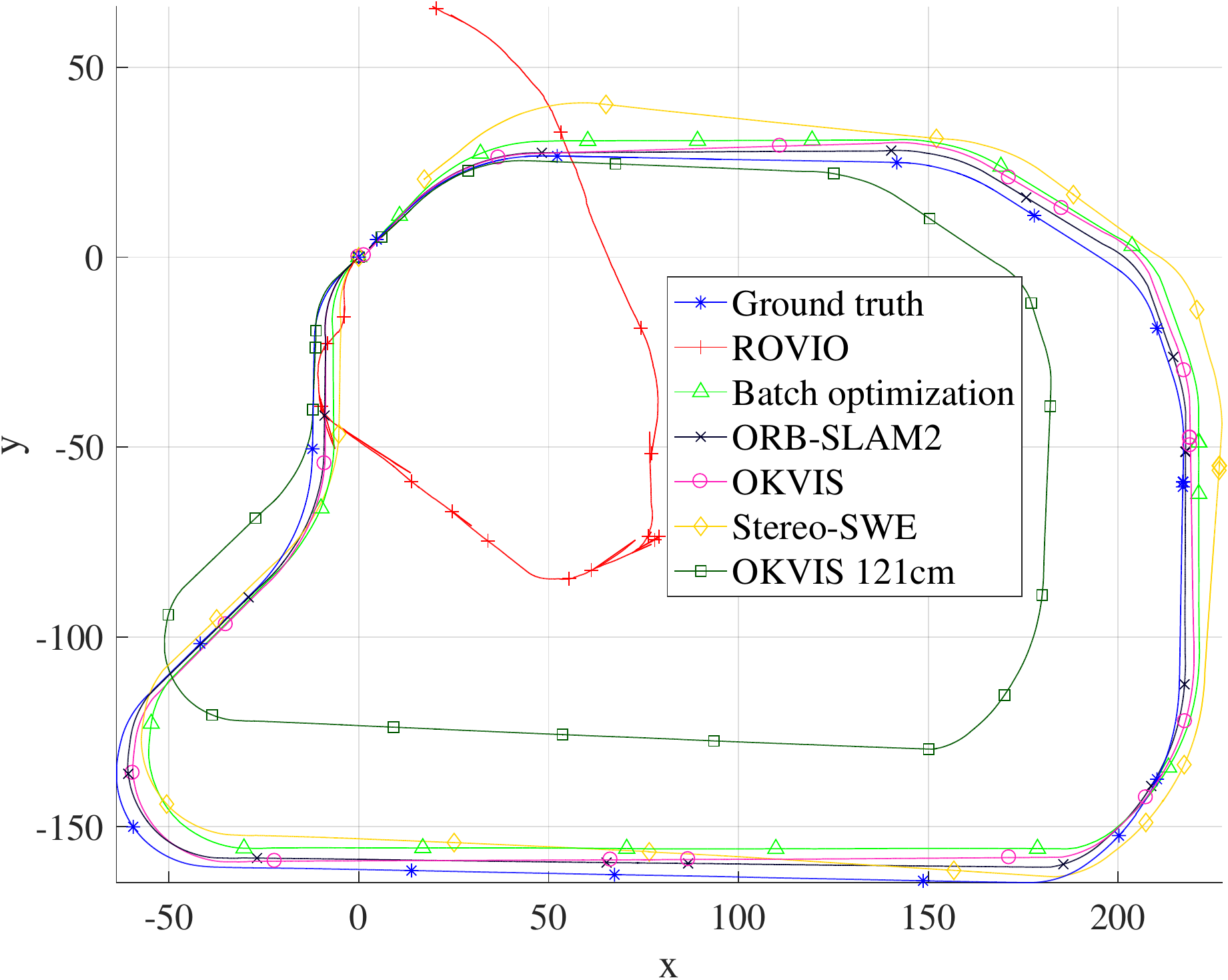}
\caption{Path alignment of trajectory estimations of the different motion estimation pipelines on \texttt{trajectory1}. If not stated otherwise, the $\unit[31]{cm}$ baseline is displayed. Due to the unrecoverable resets of the estimator, \textit{VINS-Mono} is omitted here.} \label{fig:aligned_paths_betriebshof}
\end{figure}

\begin{figure} 
\centering
\includegraphics[width=0.9\columnwidth]{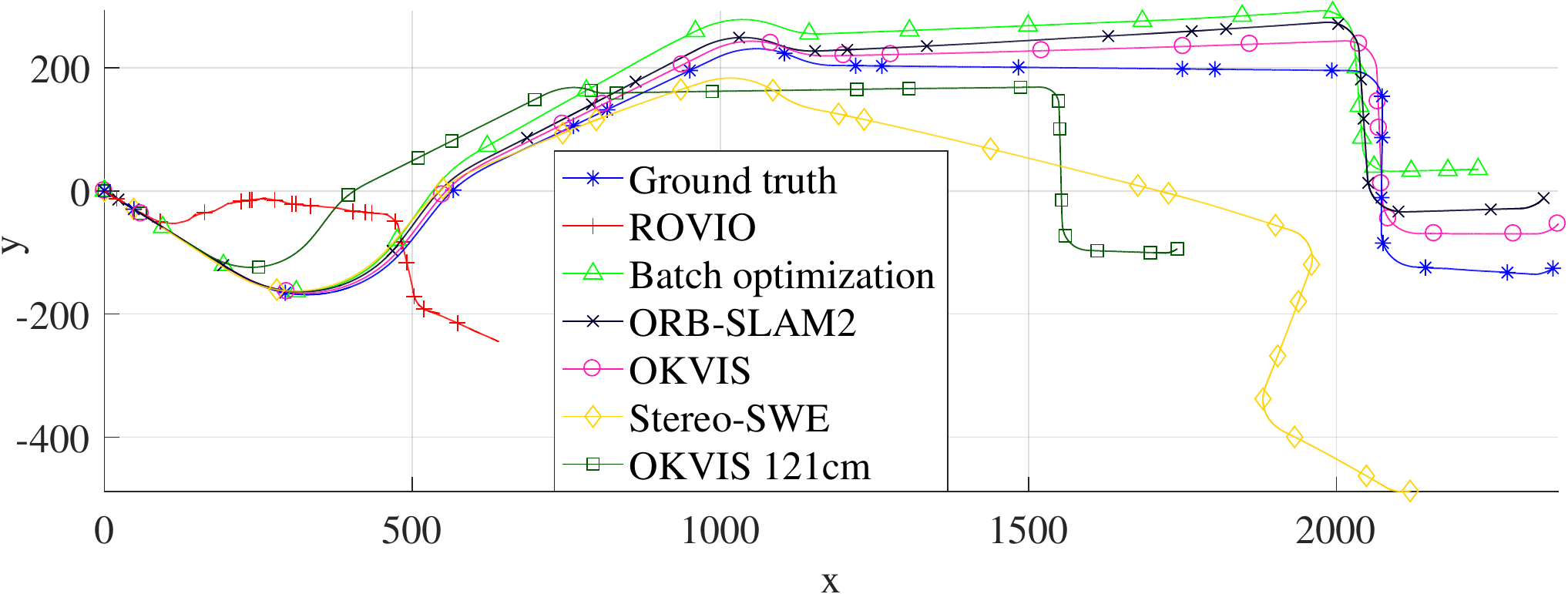}
\caption{Path alignment of trajectory estimations of the different motion estimation pipelines on \texttt{trajectory2}.  If not stated otherwise, the $\unit[31]{cm}$ baseline is displayed. Due to the unrecoverable resets of the estimator, \textit{VINS-Mono} is omitted here.} \label{fig:aligned_paths_outside}
\end{figure}

\autoref{fig:aligned_paths_betriebshof} and \autoref{fig:aligned_paths_outside} show aligned paths of the different estimated trajectories to the ground truth. 

\begin{figure} 
\centering
\includegraphics[width=\columnwidth]{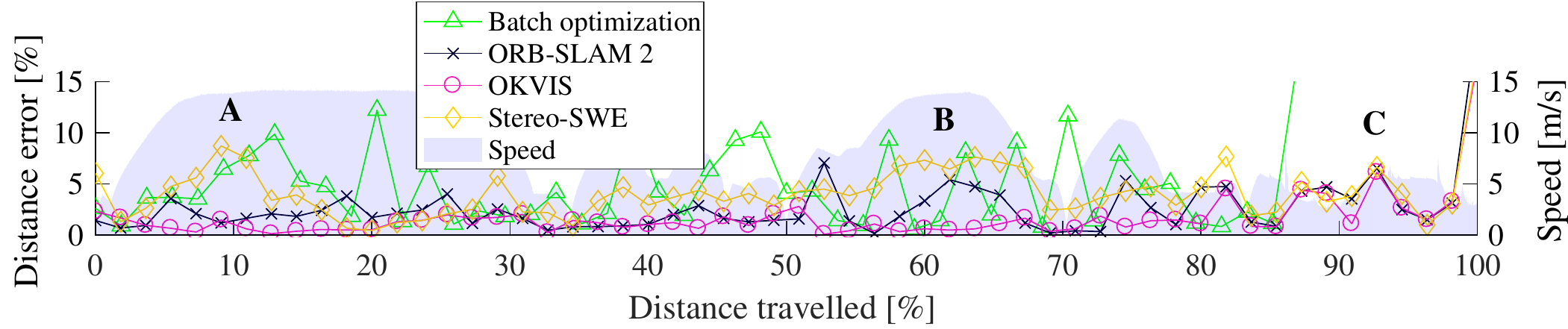} \\
\vspace{0.3cm}
\includegraphics[width=\columnwidth]{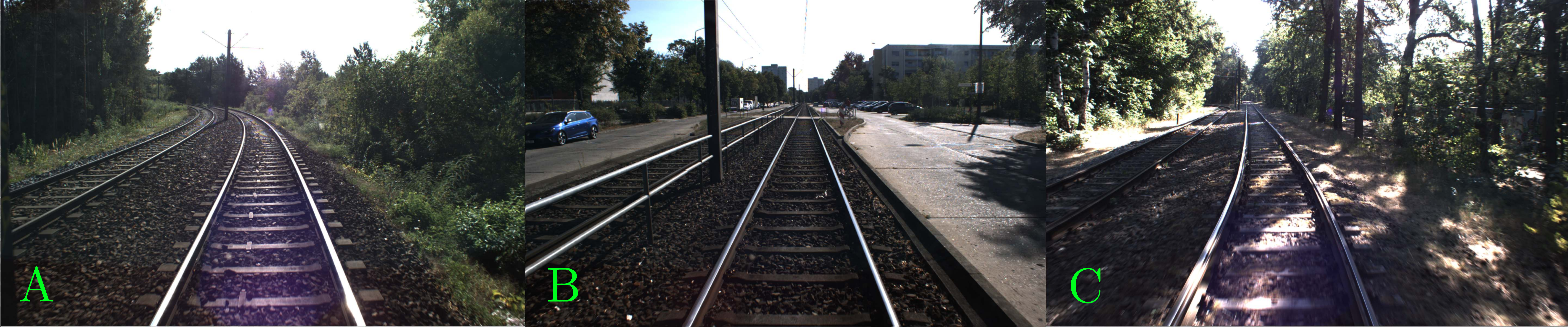}
\caption{Top: Errors of the best performing pipelines during \texttt{trajectory2} with a $\unit[31]{cm}$ baseline. The letters indicate selected challenging scenarios discussed in Section \ref{sec:discussion_and_results}. Bottom: Camera images of the challenging scenarios.} \label{fig:error_vs_traj}
\end{figure}

\subsubsection{Estimator performance}
Using the $\unit[31]{cm}$ baseline, a good calibration can be ensured. \autoref{tab:result_estimator_performance} shows the evaluation results comparing the different estimation pipelines.

Both \textit{ROVIO} and \textit{VINS-Mono} fail to work properly for rail applications.
Due to a very constrained motion and frequent constant velocity scenarios, the \ac{imu} biases cannot be estimated correctly locally. 
This results in significant scale drift, especially visible in \texttt{trajectory2} with longer sections of constant velocity.
In addition, unobservable biases lead to inconsistent estimator state and the need to re-initialize multiple times during the trajectory.
While \textit{ROVIO} can partly recover from such a reset, \textit{VINS-Mono} cannot, resulting in a somehow unfair comparison to the others.
However, \textit{ROVIO} also shows bad performance as it is highly dependent on a good knowledge of the \ac{imu} biases.

Using \textit{ROVIOLI} to build a map and \textit{maplab} \citep{Schneider2017} to globally batch bundle-adjust the maps, the scale and \ac{imu} biases can partially be recovered.
This suggests that in both \ac{ekf} and sliding-window approach, the bias estimation problem suffers significantly from linearization issues if there is not enough axis excitement in the window.
A further hint for this is the improved performance of \textit{VINS-Mono} compared to \textit{ROVIO} \reviewchanges{\sout{because of the difference in window size} which, among other possible causes, could be due to the difference in window size}.

In comparison, when using stereo constraints, metric scale can be estimated correctly during the whole trajectory.
On \texttt{trajectory1}, both \textit{OKVIS} and \textit{ORB-SLAM2} show very similar performance.
\texttt{Trajectory2} is more challenging \reviewchanges{\sout{with}due to} faster motion, \reviewchanges{more challening lighting conditions} and more dynamic objects such as cars, pedestrians and other trams in the scene.
There, \textit{OKVIS} is able to outperform \textit{ORB-SLAM2} in most cases.
\reviewchanges{This is especially visible in the \ac{rmse} in \autoref{tab:result_estimator_performance} which suggests that \textit{OKVIS} also has a higher robustness compared to \textit{ORB-SLAM2}.}
The complementing sensor modalities show benefits for motion estimation, \reviewchanges{\sout{especially}most prominently} in dynamic environments and at higher speeds.
For \textit{Stereo-SWE}, using the depth as an independent part in the optimization problem does not show an increase in accuracy.
Also, it has a drawback of increasing in tuning parameter number, which is the weighting factor between depth errors and re-projection and inertial errors.
This increases the tuning effort.

The distance errors along \texttt{trajectory2} for the best four performing estimators are shown in \autoref{fig:error_vs_traj}. 
Three distinctive challenging scenarios $\mathcal{A}$, $\mathcal{B}$ and $\mathcal{C}$ can be identified.
They are also indicated in \autoref{fig:outside}.
These challenging scenarios give evidence to the difference in estimator performances,  and are summarized in \autoref{tab:results_challenging_scenarios}.
\begin{table*}[t]
\centering
\caption{Result overview of median errors (distance in $\unit{\%}$ / heading in $\unitfrac{\deg}{m}$) changing the baselines. The best pipeline in the respective scenario is emphasized in bold letters. $^4$ Lo\reviewchanges{\sout{o}}ses track at higher speeds after about $\unit[7-8]{\%}$ of the trajectory.} \label{tab:result_baseline}
\begin{tabular}{c|l||ccc|ccc}
&Segment length& \multicolumn{3}{c|}{$\unit[10]{m}$} & \multicolumn{3}{c}{$\unit[50]{m}$}\\
&Baseline& $\unit[31]{cm}$ & $\unit[71]{cm}$ & $\unit[120]{cm}$ & $\unit[31]{cm}$ & $\unit[71]{cm}$ & $\unit[120]{cm}$ \\
\hline \hline
%
\parbox[t]{2mm}{\multirow{3}{*}{\rotatebox[origin=c]{90}{\texttt{Traj1}}}}  &ORB-SLAM2 \citep{Mur-Artal2017} & $\mathbf{2.138} \,/\, \mathbf{0.0204} $ & $\mathbf{2.701} \,/\, 0.0199 $ & $30.128 \,/\, 0.1140\footnotemark[4] $ & $3.054 \,/\, 0.0093 $ & $\mathbf{5.002} \,/\, 0.0110 $ & $33.862 \,/\, 2.8459\footnotemark[4]$\\
&OKVIS \citep{Leutenegger2015} & $2.152 \,/\, 0.0219 $ & $3.077 \,/\, 0.0198 $ & $20.733 \,/\,\mathbf{0.0154} $ & $\mathbf{2.850} \,/\, \mathbf{0.0070} $ & $5.049\,/\, \mathbf{0.0055} $ & $21.184 \,/\,\mathbf{0.0075} $\\
&Stereo-SWE \citep{Hinzmann2016MonocularOptimization}& $2.845 \,/\, 0.0249$ & $4.543 \,/\, \mathbf{0.0177} $ & $\mathbf{18.340} \,/\, 0.0225 $ & $4.029\,/\, 0.0128$ & $6.649 \,/\, 0.0081 $ & $\mathbf{19.247} \,/\, 0.0206 $\\
\hline
%
\parbox[t]{2mm}{\multirow{3}{*}{\rotatebox[origin=c]{90}{\texttt{Traj2}}}} &ORB-SLAM2 \citep{Mur-Artal2017} & $1.786 \,/\, 0.0078 $ & $\mathbf{3.247} \,/\, 0.0094 $ & $32.121 \,/\, 0.0478\footnotemark[4] $ & $1.829 \,/\, \mathbf{0.0033} $ & $3.783 \,/\, 0.0049 $ & $26.772 \,/\, 0.0305 \footnotemark[4]$\\
&OKVIS \citep{Leutenegger2015} & $\mathbf{1.428} \,/\, \mathbf{0.0074} $ & $3.465 \,/\, \mathbf{0.0068} $ & $30.947 \,/\, \mathbf{0.0072} $ & $\mathbf{1.110} \,/\, 0.0038 $ & $\mathbf{3.609} \,/\, \mathbf{0.0036} $ & $29.045 \,/\, \mathbf{0.0035} $\\
&Stereo-SWE \citep{Hinzmann2016MonocularOptimization} & $3.710 \,/\, 0.0099$ & $5.621 \,/\, 0.0096 $ & $\mathbf{24.299} \,/\, 0.0097 $ & $3.840\,/\, 0.0087$ & $6.152 \,/\, 0.0060 $ & $\mathbf{25.271} \,/\, 0.0054 $\\
\end{tabular}
\end{table*}
\subsubsection{Challenging scenarios}
\begin{table}
\caption{Challenging scenarios of the motion estimation pipelines.}\label{tab:results_challenging_scenarios}
\begin{tabular}{m{1cm}||m{1.5cm}|m{1.5cm}|m{2.6cm}}
\reviewchanges{Scenario}&\makecell[l]{Affected\\ estimators}&Cause&Solution \\ \hline \hline
$\mathcal{A}$&\textit{Stereo-SWE}&High speed&Optimize feature tracking.\\ \hline
$\mathcal{B}$& \textit{ORB-SLAM2}, \textit{Stereo-SWE}&High speed \& Aliasing&Detect and neglect affected region; use IMU fusion.\\ \hline
$\mathcal{C}$& all&Lighting conditions& Shadow compensation \citep{McManus2014ShadyInvariance}; improve auto exposure \citep{Nourani-Vatani2007}.\\
\end{tabular}
\end{table}
Scenario $\mathcal{A}$ is visible approximately $\unit[10]{\%}$ of the way through the trajectory. 
There, the tram is moving with high velocity, which increases the complexity of feature tracking. 
While the optimized feature tracking of \textit{ORB-SLAM2} and \textit{OKVIS} are able to deal with this, \textit{Stereo-SWE} and \textit{ROVIOLI} (used to build the map for batch optimization), which share the same feature-matching algorithm, have trouble finding enough feature matches.

\reviewchanges{\sout{In} Around} \reviewchanges{\sout{section}scenario} $\mathcal{B}$, the tram is also moving at high speeds \reviewchanges{as shown in \autoref{fig:error_vs_traj}.}
However, in contrast to \reviewchanges{\sout{section} scenario} $\mathcal{A}$, there is no curve and the optical flow of all nearby structure is in the same direction as the repetitive patterns on the ground such as railings or railroad ties.
This leads to aliasing and wrong feature matches.
More evidence can also be found at the beginning of \texttt{trajectory2} before entering the curve.
Using the \ac{imu} as a complementing sensor modality, as in \textit{OKVIS}, seems to be beneficial in this case.
\reviewchanges{\sout{Furthermore, one could use Nyquist theory to detect and neglect affected regions.}}
\textit{Stereo-SWE} still has troubles with feature tracking as in scenario $\mathcal{A}$.
\reviewchanges{
By masking out the area in-front of the vehicle where most visual aliasing is happening, different behaviours for the estimators can be observed as shown in \autoref{fig:masked}.
While \textit{OKVIS} behaves almost the same, the increase in error in scenario $\mathcal{B}$ for \textit{ORB-SLAM2} can be reduced. 
However, by removing some of the visual information in slow sections and no other available measurement source, the level of robustness is decreased.
This can be observed after about $\unit[83]{\%}$ of the trajectory where short heavy reflections in the upper part of the image can cause huge errors.
In this scenario, one could use Nyquist theory to detect and neglect affected regions instead of neglecting static parts of the image.
The \textit{SWE} shows higher errors in scenario $\mathcal{B}$ due to the reduced visual information in an already challenging feature-matching scenario.
\begin{figure}
    \centering
    \includegraphics[width=\columnwidth]{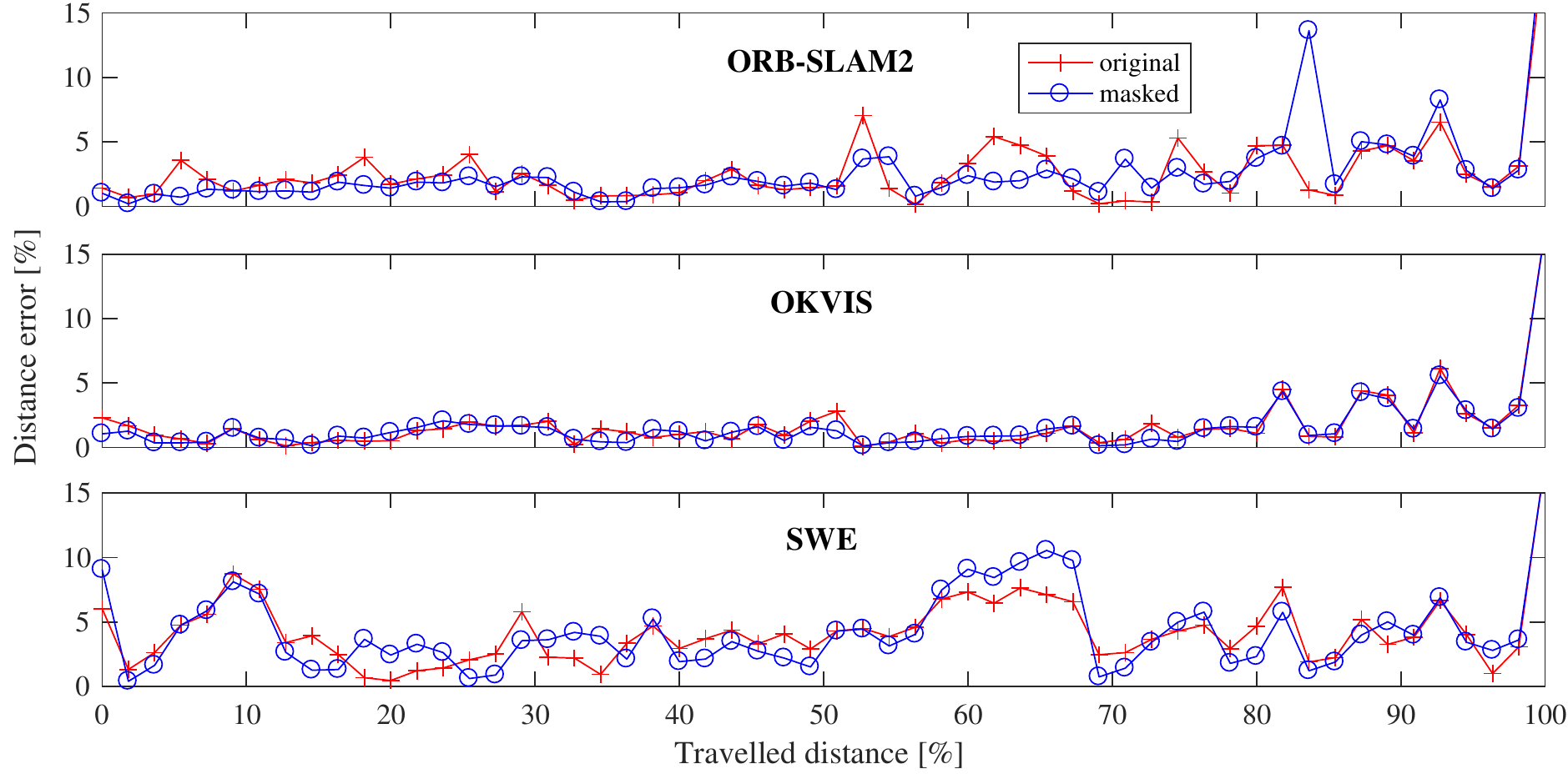}
    \caption{\reviewchanges{Distance errors using the whole image (red) and masked out visual aliasing areas (blue).}}
    \label{fig:masked}
\end{figure}
}

Finally, scenario $\mathcal{C}$ occurs in the woods at the end of the trajectory.
Fast switches from shadows to direct sunlight seem to be a challenge for all investigated pipelines.
Using improved auto-exposure control \citep{Nourani-Vatani2007} and shadow compensation \citep{McManus2014ShadyInvariance} might be beneficial.

\subsubsection{Baseline comparison}
In general, a larger baseline should provide more reliable depth information as the quantization error in the disparity is less prominent.
However, the quality of depth calculation is highly sensitive to a good calibration.
Using conventional methods, calibration is much more challenging using a higher baseline as it is harder to guarantee mutual observations of the calibration target for both cameras.
This results in a degrading calibration quality with higher baselines.
\autoref{tab:result_baseline} summarizes evaluation results using the stereo visual(-inertial) pipelines with different baselines.

For the stereo algorithms shown in the scenarios of \texttt{trajectory1} and \texttt{trajectory2}, it seems more important to have an accurate calibration using a small baseline than to be able to reliably incorporate further away landmarks.
Using \ac{imu} measurements, motion estimation is still possible up to a fixed scale error, visible in \autoref{fig:aligned_paths_betriebshof} and \autoref{fig:aligned_paths_outside}, while \textit{ORB-SLAM2} l\reviewchanges{\sout{o}}oses track at higher speeds.
In order to benefit from the advantages of higher baselines, improved calibration procedures such as online calibration \citep{Schneider2017Visual-inertialSegments} could have a high benefit.

\section{Conclusions}
Being able to accurately localize a rail vehicle has a high potential to improve infrastructure efficiency.
In a real-world application, high safety levels are typically achieved using redundant systems.
This paper studies the contribution visual systems can provide to getting closer to robust high accuracy odometry.
We did an in-depth experimental comparison using real-world rail datasets of several state-of-the-art visual-aided odometry pipelines.

It was observed that the monocular visual-inertial odometry methods \textit{ROVIO} and \textit{VINS-Mono} experience severe scale drift and are not able to keep a consistent estimator state due to locally unobservable \ac{imu} biases.
Using stereo vision, accurate motion estimation is achievable, especially using the stereo visual-inertial pipeline \textit{OKVIS}.
Specific challenging scenarios for the individual pipelines can result from high speeds, aliasing with repetitive patterns on the ground, and challenging lighting conditions.

In conclusion, even without enforcing specific motions, visual-aided odometry can achieve high accuracies for rail vehicles, but is not reliable enough for use in isolation for safety critical applications.
However, in combination with other odometry and localization methods such as \ac{gnss}, wheel odometry or ground radars, vision can complement for failure cases of other sensors.
\reviewchanges{Motion constraints can be incorporated either as a separate part in the estimation pipeline or directly into the investigated algorithms using a motion model in the propagation state for \ac{ekf} based algorithms or additional motion model error term in optimization based algorithms.}
High-speed scenarios will require higher frame-rates to ensure feature tracking and a larger baseline for reliable depth estimation of unblurred landmarks implying calibration challenges.
\reviewchanges{
Furthermore, all tested datasets are recorded during nice weather.
Like most visual pipelines, the investigated approaches will \finalsubmission{\sout{highly}}suffer from limited visibility.
However, using cameras with extended spectral band sensitivity such as \ac{lwir} shows potential to enable also good performance in bad weather conditions \cite{Pinchon2016All-weatherBand}.
}

\section*{Acknowledgement}
This work was generously supported by Siemens Mobility, Germany.
The authors would like to thank Andreas Pfrunder for his help in initial data collections and evaluations.

\bibliographystyle{IEEEtran}
\bibliography{add_bib.bib,IEEEabrv,Mendeley_Florian_PhD.bib}

\begin{acronym}
\acro{imu}[IMU]{inertial measurement unit}
\acro{etcs}[ETCS]{European Train Control System}

\acro{gps}[GPS]{Global Position System}
\acro{gnss}[GNSS]{global navigation satellite system}
\acro{asl}[ASL]{Autonomous Systems Lab}
\acro{rtk}[RTK]{real time kinematics}
\acro{slam}[SLAM]{Simultaneous Localization and Mapping}
\acro{etsc}[ETSC]{European Train Security Council}
\acro{dvs}[DVS]{Dynamic Vision Sensor}
\acro{zvv}[ZVV]{Z\"urich Verkehrsverein}
\acro{ekf}[EKF]{extended Kalman filter}
\acro{vi}[VI]{visual-inertial}
\acro{vio}[VIO]{visual-inertial odometry}
\acro{vo}[VO]{visual odometry}
\acro{lssvm}[LSSVM]{least squares support vector machine}
\acro{pf}[PF]{particle filter}
\acro{dso}[DSO]{direct sparse odometry}
\acro{mcu}[MCU]{micro controller unit}
\acro{gtsam}[GTSAM]{Georgia Tech Smoothing and Mapping library}
\acro{swe}[SWE]{Sliding Window Estimator}
\acro{uav}[UAV]{unmanned aerial vehicle}
\acro{dso}[DSO]{Direct Sparse Odometry}
\acro{rmse}[RMSE]{root mean squared error}
\acro{lwir}[LWIR]{long-wave infrared}

\end{acronym}

\end{document}